\documentclass[10pt,twocolumn,letterpaper]{article}

\usepackage{cvpr}
\usepackage{times}
\usepackage{epsfig}
\usepackage{graphicx}
\usepackage{amsmath}
\usepackage{amssymb}
\usepackage{multirow}
\usepackage{booktabs}

\newcommand{\finalcopy}{\cvprfinalcopy}
\usepackage{tikz}
\usetikzlibrary{arrows,shapes,calc,matrix,fit,backgrounds,positioning}
\usepackage{pgfplots}
\usepackage{pgfplotstable}
\pgfplotsset{compat=1.9}

\usepackage{xstring}

\usepgfplotslibrary{external}
\newcommand{\extfig}[2]{\tikzsetnextfilename{fig/extern/#1}{#2}}

\IfBeginWith*{\jobname}{fig/extern/}{\finalcopy}{}

\newcommand{\leg}[1]{\addlegendentry{#1}}

\tikzset{every mark/.append style={solid}}
\pgfplotsset{%smooth,
	grid=both, width=\columnwidth, try min ticks=5,
	every axis/.append style={font=\scriptsize},
	every axis plot/.append style={thick,mark=none,mark size=1.2,tension=0.18},
	legend cell align=left, legend style={fill opacity=0.8},
}

\pgfplotsset{
	dash/.style={mark=o,dashed,opacity=0.7},
	dott/.style={mark=o,dotted,opacity=0.7},
}

\tikzstyle{tight} = [inner sep=0pt,outer sep=0pt]

\usepackage{tabularx}

\usepackage[pagebackref=true,breaklinks=true,letterpaper=true,colorlinks,bookmarks=false]{hyperref}

\cvprfinalcopy % *** Uncomment this line for the final submission

\def\cvprPaperID{4215} % *** Enter the CVPR Paper ID here

\ifcvprfinal\fi
\begin{document}

%%%%%%%%% TITLE
\title{Dense Classification and Implanting for Few-Shot Learning}

\author{Yann Lifchitz$^{1,2}$
\ \ \ \
Yannis Avrithis$^1$
\ \ \ \
Sylvaine Picard$^2$
\ \ \ \
Andrei Bursuc$^3$
\\
{\fontsize{11}{13}\selectfont$^1$Univ Rennes, Inria, CNRS, IRISA\ \ \ \ \ \ \ $^2$Safran\ \ \ \ \ \ \ $^3$valeo.ai}
}

\maketitle

\newcommand{\head}[1]{{\smallskip\noindent\bf #1}}
\newcommand{\alert}[1]{{\color{red}{#1}}}
\newcommand{\eq}[1]{(\ref{eq:#1})\xspace}

\newcommand{\red}[1]{{\color{red}{#1}}}
\newcommand{\blue}[1]{{\color{blue}{#1}}}
\newcommand{\green}[1]{{\color{green}{#1}}}
\newcommand{\gray}[1]{{\color{gray}{#1}}}

%--------------------------------------------------------------------

\newcommand{\tran}{^\top}
\newcommand{\mtran}{^{-\top}}
\newcommand{\zcol}{\mathbf{0}}
\newcommand{\zrow}{\zcol\tran}

\newcommand{\ind}{\mathbbm{1}}
\newcommand{\expect}{\mathbb{E}}
\newcommand{\nat}{\mathbb{N}}
\newcommand{\zahl}{\mathbb{Z}}
\newcommand{\real}{\mathbb{R}}
\newcommand{\proj}{\mathbb{P}}
\newcommand{\prob}{\mathbf{Pr}}

\newcommand{\mif}{\textrm{if }}
\newcommand{\minimize}{\textrm{minimize }}
\newcommand{\maximize}{\textrm{maximize }}
\newcommand{\st}{\textrm{subject to }}

\newcommand{\id}{\operatorname{id}}
\newcommand{\const}{\operatorname{const}}
\newcommand{\sgn}{\operatorname{sgn}}
\newcommand{\var}{\operatorname{Var}}
\newcommand{\mean}{\operatorname{mean}}
\newcommand{\trace}{\operatorname{tr}}
\newcommand{\diag}{\operatorname{diag}}
\newcommand{\vect}{\operatorname{vec}}
\newcommand{\cov}{\operatorname{cov}}

\newcommand{\softmax}{\operatorname{softmax}}
\newcommand{\clip}{\operatorname{clip}}

\newcommand{\defn}{\mathrel{:=}}
\newcommand{\peq}{\mathrel{+\!=}}
\newcommand{\meq}{\mathrel{-\!=}}

\newcommand{\floor}[1]{\left\lfloor{#1}\right\rfloor}
\newcommand{\ceil}[1]{\left\lceil{#1}\right\rceil}
\newcommand{\inner}[1]{\left\langle{#1}\right\rangle}
\newcommand{\norm}[1]{\left\|{#1}\right\|}
\newcommand{\frob}[1]{\norm{#1}_F}
\newcommand{\card}[1]{\left|{#1}\right|\xspace}
\newcommand{\diff}{\mathrm{d}}
\newcommand{\der}[3][]{\frac{d^{#1}#2}{d#3^{#1}}}
\newcommand{\pder}[3][]{\frac{\partial^{#1}{#2}}{\partial{#3^{#1}}}}
\newcommand{\ipder}[3][]{\partial^{#1}{#2}/\partial{#3^{#1}}}
\newcommand{\dder}[3]{\frac{\partial^2{#1}}{\partial{#2}\partial{#3}}}

\newcommand{\wb}[1]{\overline{#1}}
\newcommand{\wt}[1]{\widetilde{#1}}

\def\xssp{\hspace{1pt}}
\def\ssp{\hspace{3pt}}
\def\msp{\hspace{5pt}}
\def\lsp{\hspace{12pt}}

\newcommand{\cA}{\mathcal{A}}
\newcommand{\cB}{\mathcal{B}}
\newcommand{\cC}{\mathcal{C}}
\newcommand{\cD}{\mathcal{D}}
\newcommand{\cE}{\mathcal{E}}
\newcommand{\cF}{\mathcal{F}}
\newcommand{\cG}{\mathcal{G}}
\newcommand{\cH}{\mathcal{H}}
\newcommand{\cI}{\mathcal{I}}
\newcommand{\cJ}{\mathcal{J}}
\newcommand{\cK}{\mathcal{K}}
\newcommand{\cL}{\mathcal{L}}
\newcommand{\cM}{\mathcal{M}}
\newcommand{\cN}{\mathcal{N}}
\newcommand{\cO}{\mathcal{O}}
\newcommand{\cP}{\mathcal{P}}
\newcommand{\cQ}{\mathcal{Q}}
\newcommand{\cR}{\mathcal{R}}
\newcommand{\cS}{\mathcal{S}}
\newcommand{\cT}{\mathcal{T}}
\newcommand{\cU}{\mathcal{U}}
\newcommand{\cV}{\mathcal{V}}
\newcommand{\cW}{\mathcal{W}}
\newcommand{\cX}{\mathcal{X}}
\newcommand{\cY}{\mathcal{Y}}
\newcommand{\cZ}{\mathcal{Z}}

\newcommand{\vA}{\mathbf{A}}
\newcommand{\vB}{\mathbf{B}}
\newcommand{\vC}{\mathbf{C}}
\newcommand{\vD}{\mathbf{D}}
\newcommand{\vE}{\mathbf{E}}
\newcommand{\vF}{\mathbf{F}}
\newcommand{\vG}{\mathbf{G}}
\newcommand{\vH}{\mathbf{H}}
\newcommand{\vI}{\mathbf{I}}
\newcommand{\vJ}{\mathbf{J}}
\newcommand{\vK}{\mathbf{K}}
\newcommand{\vL}{\mathbf{L}}
\newcommand{\vM}{\mathbf{M}}
\newcommand{\vN}{\mathbf{N}}
\newcommand{\vO}{\mathbf{O}}
\newcommand{\vP}{\mathbf{P}}
\newcommand{\vQ}{\mathbf{Q}}
\newcommand{\vR}{\mathbf{R}}
\newcommand{\vS}{\mathbf{S}}
\newcommand{\vT}{\mathbf{T}}
\newcommand{\vU}{\mathbf{U}}
\newcommand{\vV}{\mathbf{V}}
\newcommand{\vW}{\mathbf{W}}
\newcommand{\vX}{\mathbf{X}}
\newcommand{\vY}{\mathbf{Y}}
\newcommand{\vZ}{\mathbf{Z}}

\newcommand{\va}{\mathbf{a}}
\newcommand{\vb}{\mathbf{b}}
\newcommand{\vc}{\mathbf{c}}
\newcommand{\vd}{\mathbf{d}}
\newcommand{\ve}{\mathbf{e}}
\newcommand{\vf}{\mathbf{f}}
\newcommand{\vg}{\mathbf{g}}
\newcommand{\vh}{\mathbf{h}}
\newcommand{\vi}{\mathbf{i}}
\newcommand{\vj}{\mathbf{j}}
\newcommand{\vk}{\mathbf{k}}
\newcommand{\vl}{\mathbf{l}}
\newcommand{\vm}{\mathbf{m}}
\newcommand{\vn}{\mathbf{n}}
\newcommand{\vo}{\mathbf{o}}
\newcommand{\vp}{\mathbf{p}}
\newcommand{\vq}{\mathbf{q}}
\newcommand{\vr}{\mathbf{r}}
\newcommand{\Vs}{\mathbf{s}}
\newcommand{\vt}{\mathbf{t}}
\newcommand{\vu}{\mathbf{u}}
\newcommand{\vv}{\mathbf{v}}
\newcommand{\vw}{\mathbf{w}}
\newcommand{\vx}{\mathbf{x}}
\newcommand{\vy}{\mathbf{y}}
\newcommand{\vz}{\mathbf{z}}

\newcommand{\vone}{\mathbf{1}}
\newcommand{\vzero}{\mathbf{0}}

\newcommand{\valpha}{{\boldsymbol{\alpha}}}
\newcommand{\vbeta}{{\boldsymbol{\beta}}}
\newcommand{\vgamma}{{\boldsymbol{\gamma}}}
\newcommand{\vdelta}{{\boldsymbol{\delta}}}
\newcommand{\vepsilon}{{\boldsymbol{\epsilon}}}
\newcommand{\vzeta}{{\boldsymbol{\zeta}}}
\newcommand{\veta}{{\boldsymbol{\eta}}}
\newcommand{\vtheta}{{\boldsymbol{\theta}}}
\newcommand{\viota}{{\boldsymbol{\iota}}}
\newcommand{\vkappa}{{\boldsymbol{\kappa}}}
\newcommand{\vlambda}{{\boldsymbol{\lambda}}}
\newcommand{\vmu}{{\boldsymbol{\mu}}}
\newcommand{\vnu}{{\boldsymbol{\nu}}}
\newcommand{\vxi}{{\boldsymbol{\xi}}}
\newcommand{\vomikron}{{\boldsymbol{\omikron}}}
\newcommand{\vpi}{{\boldsymbol{\pi}}}
\newcommand{\vrho}{{\boldsymbol{\rho}}}
\newcommand{\vsigma}{{\boldsymbol{\sigma}}}
\newcommand{\vtau}{{\boldsymbol{\tau}}}
\newcommand{\vupsilon}{{\boldsymbol{\upsilon}}}
\newcommand{\vphi}{{\boldsymbol{\phi}}}
\newcommand{\vchi}{{\boldsymbol{\chi}}}
\newcommand{\vpsi}{{\boldsymbol{\psi}}}
\newcommand{\vomega}{{\boldsymbol{\omega}}}

\newcommand{\rLambda}{\mathrm{\Lambda}}
\newcommand{\rSigma}{\mathrm{\Sigma}}

%--------------------------------------------------------------------
% Add a period to the end of an abbreviation unless there's one
% already, then \xspace.
\makeatletter
\DeclareRobustCommand\onedot{\futurelet\@let@token\@onedot}
\def\@onedot{\ifx\@let@token.\else.\null\fi\xspace}
\def\eg{\emph{e.g}\onedot} \def\Eg{\emph{E.g}\onedot}
\def\ie{\emph{i.e}\onedot} \def\Ie{\emph{I.e}\onedot}
\def\cf{\emph{cf}\onedot} \def\Cf{\emph{Cf}\onedot}
\def\etc{\emph{etc}\onedot} \def\vs{\emph{vs}\onedot}
\def\wrt{w.r.t\onedot} \def\dof{d.o.f\onedot}
\def\etal{\emph{et al}\onedot}
\makeatother
\newcommand{\secref}[1]{\S\ref{#1}}
\newcommand{\classify}{densely classify\xspace}
\newcommand{\classification}{dense classification\xspace}
\newcommand{\Classification}{Dense classification\xspace}
\newcommand{\short}{DC\xspace}

\newcommand{\Implanting}{Implanting\xspace}
\newcommand{\implanting}{implanting\xspace}
\newcommand{\Implantingshort}{IMP\xspace}

\newcommand{\AVGtraining}{Global average pooling\xspace}
\newcommand{\aVGtraining}{global average pooling\xspace}
\newcommand{\AVGpoolshort}{GAP\xspace}
\newcommand{\MAXpoolshort}{GMP\xspace}

\newcommand{\kay}{k}

\newcommand{\new}[1]{#1'}

\newcommand{\nC}{\new{C}}
\newcommand{\nN}{\new{N}}
\newcommand{\nW}{\new{W}}
\newcommand{\nX}{\new{X}}
\newcommand{\nY}{\new{Y}}

\newcommand{\nva}{\new{\va}}
\newcommand{\nc}{\new{c}}
\newcommand{\nn}{\new{n}}
\newcommand{\nvx}{\new{\vx}}
\newcommand{\ny}{\new{y}}

\newcommand{\ntheta}{\new{\theta}}

%%%%%%%%% ABSTRACT
\begin{abstract}
   Training deep neural networks from few examples is a highly challenging and key problem for many computer vision tasks. In this context, we are targeting knowledge transfer from a set with abundant data to other sets with few available examples. We propose two simple and effective solutions: (i) \emph{\classification} over feature maps, which for the first time studies local activations in the domain of few-shot learning, and (ii) \emph{implanting}, that is, attaching new neurons to a previously trained network to learn new, task-specific features. On miniImageNet, we improve the prior state-of-the-art on few-shot classification, \ie, we achieve 62.5\%, 79.8\% and 83.8\% on 5-way 1-shot, 5-shot and 10-shot settings respectively.
\end{abstract}

\vspace{-0.3cm}
\section{Introduction}
\label{sec:intro}

Current state of the art on image classification~\cite{simonyan2014, he2016, huang2017}, object detection~\cite{liu2016b, redmon2016, he2017}, semantic segmentation~\cite{yu2017, chen2017, liu2018}, and practically most tasks with some degree of learning involved, rely on deep neural networks.
Those are powerful high-capacity models with trainable parameters ranging from millions to tens of millions, which require vast amounts of annotated data to fit. When such data is plentiful, supervised learning is the solution of choice.

Tasks and classes with limited available data, \ie from the long-tail~\cite{wang2017}, are highly problematic for this type of approaches. The performance of deep neural networks poses several challenges in the low-data regime, in particular in terms of overfitting and generalization. The subject of few-shot learning is to learn to recognize previously unseen classes with very few annotated examples. This is not a new problem~\cite{fei2006}, yet there is a recent resurgence in interest through \emph{meta-learning}~\cite{koch2015, vinyals2016, santoro2016, bertinetto2016, finn2017} inspired by early work in \emph{learning-to-learn}~\cite{thrun1998, hochreiter2001}.

In meta-learning settings, even when there is single large training set with a fixed number of class, it is treated as a collection of datasets of different classes, where each class has a few annotated examples. This is done so that both \emph{meta-learning} and \emph{meta-testing} are performed in a similar manner~\cite{vinyals2016, santoro2016, finn2017}. However this choice does not always come with best performance. We argue that a simple conventional pipeline using all available classes and data with a \emph{parametric classifier} is effective and appealing.

Most few-shot learning approaches do not deal explicitly with spatial information since feature maps are usually flattened or pooled before the classification layer. We show that performing a \emph{\classification} over feature maps leads to more precise classification and consistently improves performance on standard benchmarks.

While \emph{incremental learning} touches similar aspects with few-shot learning by learning to adapt to new tasks using the same network~\cite{mallya2018a, mallya2018b} or extending an existing network with new layers and parameters for each new task~\cite{rusu2016}, few of these ideas have been adopted in few shot learning. The main impediment is the reduced number of training examples which make it difficult to properly define a new task. We propose a solution for leveraging incremental learning ideas for few-shot learning.

\head{Contributions:} We present the following contributions. First, we propose a simple extension for few-shot learning pipelines consisting of \emph{\classification} over feature maps. Through localized supervision, it enables reaping additional knowledge from the limited training data. Second, we introduce neural \emph{implants}, which are layers attached to an already trained network, enabling it to quickly adapt to new tasks with few examples. Both are easy to implement and show consistent performance gains.

\section{Problem formulation and background}
\label{sec:background}

\head{Problem formulation.} We are given a collection of \emph{training} examples $X \defn (\vx_1,\dots,\vx_n)$ with each $\vx_i \in \cX$, and corresponding labels $Y \defn (y_1,\dots,y_n)$ with each $y_i \in C$, where $C \defn [c]$\footnote{We use the notation $[i] \defn \{1,\dots,i\}$ for $i \in \nat$.} is a set of \emph{base classes}. On this training data we are allowed to learn a representation of the domain $\cX$ such that we can solve new tasks. This representation learning we shall call \emph{stage 1}.

In few-shot learning, one new task is that we are given a collection of few \emph{support} examples $\nX \defn (\nvx_1,\dots,\nvx_{\nn})$ with each $\nvx_i \in \cX$, and corresponding labels $\nY \defn (\ny_1,\dots,\ny_{\nn})$ with each $\ny_i \in \nC$, where $\nC \defn [\nc]$ is a set of \emph{novel classes} disjoint from $C$ and $\nn \ll n$; with this new data, the objective is to learn a classifier that maps a new \emph{query} example from $\cX$ to a label prediction in $\nC$. The latter classifier learning, which does not exclude continuing the representation learning, we shall call \emph{stage 2}.

Classification is called \emph{$\nc$-way} where $\nc$ is the number of novel classes; in case there is a fixed number $\kay$ of support examples per novel class, it is called \emph{$\kay$-shot}. As in standard classification, there is typically a collection of queries for evaluation of each task. Few-shot learning is typically evaluated on a large number of new tasks, with queries and support examples randomly sampled from $(\nX,\nY)$.

\head{Network model.} We consider a model that is conceptually composed of two parts: an embedding network and a classifier. The \emph{embedding network} $\phi_\theta: \cX \to \real^{r \times d}$ maps the input to an embedding, where $\theta$ denotes its parameters. Since we shall be studying the spatial properties of the input, the embedding is not a vector but rather a tensor, where $r$ represents the spatial dimensions and $d$ the feature dimensions. For a 2d input image and a convolutional network for instance, the embedding is a 3d tensor in $\real^{w \times h \times d}$ taken as the activation of the last convolutional layer, where $r = w \times h$ is its spatial resolution. The embedding can still be a vector in the special case $r=1$.

The \emph{classifier network} can be of any form and depends on the particular model, but it is applied on top of $\phi_\theta$ and its output represents confidence over $c$ (resp. $\nc$) base (resp. novel) classes. If we denote by $f_\theta: \cX \to R^c$ (resp. $R^{\nc}$) the \emph{network function} mapping the input to class confidence, then a prediction for input $x \in \cX$ is made by assigning the label of maximum confidence, $\arg\max_i f_\theta^i(x)$\footnote{Given vector $\vx \in \real^m$, $x^i$ denotes the $i$-th element of $\vx$. Similarly for $f: A \to \real^m$, $f^i(a)$ denotes the $i$-the element of $f(a)$ for $a \in A$.}.

\head{Prototypical networks.} Snell \etal~\cite{snell2017} introduce a simple classifier for novel classes that computes a single prototype per class and then classifies a query to the nearest prototype. More formally, given $\nX,\nY$ and an index set
$S \subset \nN \defn [\nn]$,
% $I \subset \nN \defn [\nn]$,
let the set
$S_j \defn \{ i \in S: \ny_i = j \}$
% $\nN_j \defn \{ i \in I: \ny_i = j \}$
index the support examples in
$S$
% $I$
labeled in class $j$. The \emph{prototype} of class $j$ is given by the average of those examples
% \vspace{-0.07cm}
\begin{align}
	\vp_j = \frac{1}{|S_j|} \sum_{i \in S_j} \phi_\theta(\nvx_i)
% 	\vp_j = \frac{1}{|\nN_j|} \sum_{i \in \nN_j} \phi_\theta(\nvx_i)
\label{eq:proto}
\end{align}
for $j \in \nC$. Then, the network function is defined as\footnote{We define $[e(i)]_{i=1}^n \defn (e(1),\dots,e(n))$ for $n \in \nat$ and any expression $e(i)$ of variable $i \in \nat$.}
\begin{align}
	f_\theta[P](\vx) \defn \vsigma \left( [s(\phi_\theta(\vx), \vp_j)]_{j=1}^{\nc} \right)
\label{eq:proto-map}
\end{align}
for $\vx \in \cX$, where
$P \defn (\vp_1,\dots,\vp_{\nc})$ and
$s$ is a similarity function that may be cosine similarity or negative squared Euclidean distance and $\vsigma: \real^m \to \real^m$ is the \emph{softmax function} defined by
\begin{align}
	\vsigma(\vx) \defn \left[ \frac{\exp(x^j)}{\sum_{i=1}^m \exp(x^i)} \right]_{j=1}^{m}
\label{eq:softmax}
\end{align}
for $\vx \in \real^m$ and $m \in \nat$.

Given a new task with support data $(\nX,\nY)$ over novel classes $\nC$ (stage 2), the full index set $\nN$ is used and computing class prototypes~\eq{proto} is the only learning to be done.

When learning from the training data $(X,Y)$ over base classes $C$ (stage 1), a number of fictitious tasks called \emph{episodes} are generated by randomly sampling a number classes from $C$ and then a number of examples in each class from $X$ with their labels from $Y$; these collections, denoted as $\nX,\nY$ respectively and of length $\nn$, are supposed to be support examples and queries of novel classes $\nC$, where labels are now available for the queries and the objective is that queries are classified correctly. The set $\nN \defn [\nn]$ is partitioned into a \emph{support set} $S \subset \nN$ and a \emph{query set} $Q \defn \nN \setminus S$. Class prototypes $P$ are computed on index set $S$ according to~\eq{proto} and the network function $f_\theta$ is defined on these prototypes by~\eq{proto-map}. The network is then trained by minimizing over $\theta$ the \emph{cost function}
\begin{align}
	J(\nX, \nY; \theta) \defn \sum_{i \in Q} \ell(f_\theta[P](\nvx_i), \ny_i)
\label{eq:proto-cost}
\end{align}
on the query set $Q$, where $\ell$ is the \emph{cross-entropy} loss

\begin{align}
	\ell(\va,y) \defn -\log a^y
\label{eq:ce}
\end{align}
for $\va \in \real^m$, $y \in [m]$ and $m \in \nat$.

\head{Learning with imprinted weights.} Qi \etal~\cite{qi2018} follow a simpler approach when learning on the training data $(X,Y)$ over base classes $C$ (stage 1). In particular, they use a fully-connected layer without bias as a \emph{parametric linear classifier} on top of the embedding function $\phi_\theta$ followed by softmax and they train in a standard supervised classification setting. More formally, let $\vw_j \in \real^{r \times d}$ be the weight parameter of class $j$ for $j \in C$. Then, similarly to~\eq{proto-map}, the network function is defined by
\begin{align}
	f_{\theta,W}(\vx) \defn \vsigma \left( [s_\tau(\phi_\theta(\vx), \vw_j)]_{j=1}^c \right)
\label{eq:print-map}
\end{align}
for $\vx \in \cX$, where
$W \defn (\vw_1,\dots,\vw_c)$ is the collection of class weights and $s_\tau$ is the \emph{scaled cosine similarity}
\begin{align}
	s_\tau(\vx,\vy) \defn \tau \inner{\hat{\vx},\hat{\vy}}
\label{eq:cosine}
\end{align}
for $\vx,\vy \in \real^{r \times d}$; $\hat{\vx} \defn \vx / \norm{\vx}$ is the $\ell_2$-normalized counterpart of $\vx$ for $\vx \in \real^{r \times d}$; $\inner{\cdot,\cdot}$ and $\norm{\cdot}$ denote Frobenius inner product and norm respectively; and $\tau \in \real^+$ is a trainable scale parameter. Then, training amounts to minimizing over $\theta,W$ the \emph{cost function}
\begin{align}
	J(X, Y; \theta, W) \defn \sum_{i=1}^n \ell(f_{\theta,W}(\vx_i), y_i).
\label{eq:print-cost}
\end{align}

Given a new task with support data ($\nX,\nY$) over novel classes $\nC$ (stage 2), class prototypes $P$ are computed on $\nN$ according to~\eq{proto} and they are \emph{imprinted} in the classifier, that is, $W$ is replaced by $\nW \defn (W,P)$. The network can now make predictions on $n+\nn$ base and novel classes. The network is then fine-tuned based on~\eq{print-cost}, which aligns the class weights $W$ with the prototypes $P$ at the cost of having to store and re-train on the entire training data $(X,Y)$.

\head{Few-shot learning without forgetting.} Gidaris and Komodakis~\cite{gidaris2018}, concurrently with~\cite{qi2018}, develop a similar model that is able to classify examples of both base and novel classes. The main difference to~\cite{qiao2018} is that only the weight parameters of the base classes are stored and not the entire training data. They use the same \emph{parametric} linear classifier as~\cite{qi2018} in both stages, and they also use episode-style training like~\cite{snell2017} in stage 2.
\section{Method}
\label{sec:method}

Given training data of base classes (stage 1), we use a parametric classifier like~\cite{qi2018,gidaris2018}, which however applies at all spatial locations rather than following flattening or pooling; a very simple idea that we call \emph{\classification} and discuss in~\secref{sec:denseclassif}. Given support data of novel classes (stage 2), we learn in episodes as in prototypical networks~\cite{snell2017}, but on the true task. As discussed in~\secref{sec:implant}, the embedding network learned in stage 1 remains fixed but new layers called \emph{implants} are trained to learn task-specific features. Finally,~\secref{sec:classify} discusses inference of novel class queries.

\subsection{\Classification}
\label{sec:denseclassif}

\begin{figure}
\centering
\input{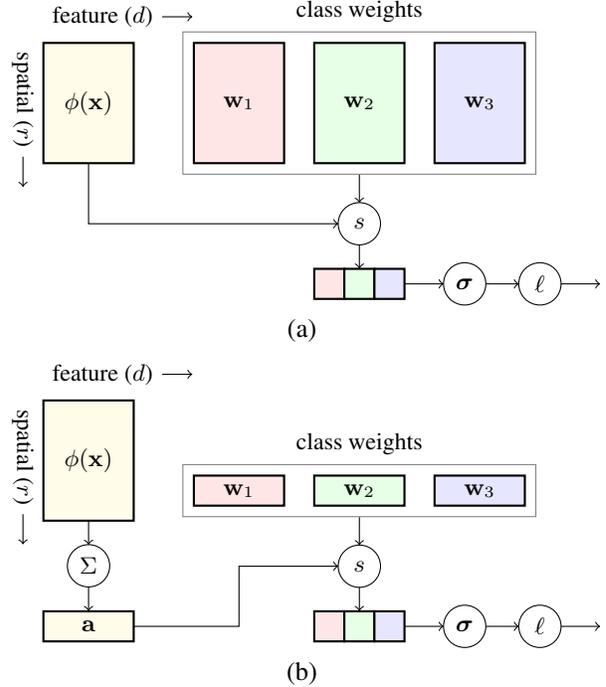}
\extfig{dc1}{\begin{tikzpicture}[pixel]
	\path  (1,0) node[box,full,ex] (ex) {} node {$\phi(\vx)$};
	\path  (6,0) node[box,full,p1] (p1) {} node {$\vw_1$};
	\path (10,0) node[box,full,p2] (p2) {} node {$\vw_2$};
	\path (14,0) node[box,full,p3] (p3) {} node {$\vw_3$};
	\node[group,fit=(p1)(p2)(p3)] (g) {};
	\node[anchor=south west] at($(ex.north west)+(0,.2)$) (ft) {feature ($d$)};
	\node[rotate=-90,anchor=north west] at(ex.north west) (sp) {spatial ($r$)};
	\draw[->] (ft.east)--+(1,0);
	\draw[->] (sp.east)--+(0,-1);
	\path (p2.south) +(0,-2) node[node] (s) {$s$};
	\path (s) ++(-1,-2) node[box,cell,p1] (s1) {}
	          ++( 1, 0) node[box,cell,p2] (s2) {}
	          ++( 1, 0) node[box,cell,p3] (s3) {};
	\node[above] at(g.north) {class weights};
	\path (s3) ++(2.5,0) node[node] (sm) {$\vsigma$}
	           ++(2.5,0) node[node] (l) {$\ell$};
	\draw[->] (ex)|-(s);
	\draw[->] (g)--(s);
	\draw[->] (s)--(s2);
	\draw[->] (s3)--(sm);
	\draw[->] (sm)--(l);
	\draw[->] (l)--+(2,0);
\end{tikzpicture}} \\ (a) \\[4pt]
\extfig{dc2}{\begin{tikzpicture}[pixel]
	\path  (1,1) node[box,full,ex] (ex) {} node {$\phi(\vx)$};
	\path  (6,0) node[box,thn,p1] (p1) {} node {$\vw_1$};
	\path (10,0) node[box,thn,p2] (p2) {} node {$\vw_2$};
	\path (14,0) node[box,thn,p3] (p3) {} node {$\vw_3$};
	\node[group,fit=(p1)(p2)(p3)] (g) {};
	\node[anchor=south west] at($(ex.north west)+(0,.2)$) (ft) {feature ($d$)};
	\node[rotate=-90,anchor=north west] at(ex.north west) (sp) {spatial ($r$)};
	\draw[->] (ft.east)--+(1,0);
	\draw[->] (sp.east)--+(0,-1);
	\path (p2.south) +(0,-2) node[node] (s) {$s$};
	\path (1,-2.5) node[node] (P) {$\Sigma$};
	\path (s) ++(-1,-2) node[box,cell,p1] (s1) {}
	          ++( 1, 0) node[box,cell,p2] (s2) {}
	          ++( 1, 0) node[box,cell,p3] (s3) {};
	\node[above] at(g.north) {class weights};
	\path let \p1=(s1) in (1,\y1) node[box,thn,ex] (po) {} node {$\va$};
	\path (s3) ++(2.5,0) node[node] (sm) {$\vsigma$}
	           ++(2.5,0) node[node] (l) {$\ell$};
	\draw[->] (ex)--(P);
	\draw[->] (P)--(po);
	\draw[->] let \p1=(p1),\p2=(po),\p3=(s) in (po)--(\x1,\y2)--(\x1,\y3)--(s);
	\draw[->] (g)--(s);
	\draw[->] (s)--(s2);
	\draw[->] (s3)--(sm);
	\draw[->] (sm)--(l);
	\draw[->] (l)--+(2,0);
\end{tikzpicture}} \\ (b)
\caption{\emph{Flattening and pooling}. Horizontal (vertical) axis represents feature (spatial) dimensions. Tensors $\vw_1,\vw_2,\vw_3$ represent class weights, and $\phi(\vx)$ the embedding of example $\vx$. An embedding is compared to class weights by similarity ($s$) and then softmax ($\vsigma$) and cross-entropy ($\ell$) follow. (a) \emph{Flattening} is equivalent to class weights having the same $r \times d$ shape as $\phi(\vx)$. (b) \emph{Global pooling}. Embedding $\phi(\vx)$ is pooled ($\Sigma$) into vector $\va \in \real^d$ before being compared to class weights, which are in $\real^d$ too.}
\label{fig:dc}
\end{figure}

\begin{figure*}
\centering
\input{fig/dc}
\extfig{dc3}{\begin{tikzpicture}[pixel]
	\path (-5,-4) node[box,full,ex] (ex) {} node {$\phi(\vx)$};
	\path (29,-4) node[op] (S) {$+$};
	\draw[->] (S)--+(2,0);
	\node[anchor=south west] at($(ex.north west)+(0,.2)$) (ft) {feature ($d$)};
	\node[rotate=-90,anchor=north west] at(ex.north west) (sp) {spatial ($r$)};
	\draw[->] (ft.east)--+(1,0);
	\draw[->] (sp.east)--+(0,-1);
	\foreach \r in {1,2,3} {
		\path  (1,-2*\r) node[box,thn,ex] (r1) {} node {$\phi^{(\r)}(\vx)$};
		\path  (6,-2*\r) node[box,thn,p1] (p1) {} node {$\vw_1$};
		\path (10,-2*\r) node[box,thn,p2] (p2) {} node {$\vw_2$};
		\path (14,-2*\r) node[box,thn,p3] (p3) {} node {$\vw_3$};
		\node[group2,fit=(p1)(p2)(p3)] (g\r) {};
		\path (p3.east) +(2,0) node[node] (s) {$s$};
		\path (s) ++(2.5,0) node[box,cell,p1] (s1) {}
		          ++(1,0) node[box,cell,p2] (s2) {}
		          ++(1,0) node[box,cell,p3] (s3) {};
		\path (s3) ++(2.5,0) node[node] (sm) {$\vsigma$}
					++(2.5,0) node[node] (l) {$\ell$};
		\draw[->] (ex)--(r1.west);
		\draw[->] (r1)--+(1,1)--($(s)+(-1,1)$)--(s);
		\draw[->] (g\r)--(s);
		\draw[->] (s)--(s1);
		\draw[->] (s3)--(sm);
		\draw[->] (sm)--(l);
		\draw[->] (l)--(S);
	}
		\node[above] at(g1.north) {class weights};
\end{tikzpicture}}
\caption{\emph{\Classification}. Notation is the same as in Figure~\ref{fig:dc}. The embedding $\va \defn \phi(\vx) \in \real^{r \times d}$ is seen as a collection of vectors $(\va^{(1)},\dots,\va^{(r)})$ in $\real^d$ (here $r=3$) with each being a vector in $\real^d$ and representing a region of the input image. Each vector is compared independently to the same class weights and the losses are added, encouraging all regions to be correctly classified.}
\label{fig:dc3}
\vspace{-2mm}
\end{figure*}

\begin{figure}
\centering
\newcommand{\img}[1]{\includegraphics[width=.24\columnwidth]{fig/saliency/#1}}
\newcommand{\ovr}[2]{\extfig{#1-#2}{\tikz{\node[tight,opacity=.7]{\img{#1_grey}};\node[tight,opacity=.45]{\img{#1_sim_#2_color}};}}}
\newcommand{\pair}[1]{\ovr{#1}{pooling}\ \ovr{#1}{slicing}}
\newcommand{\lab}[1]{\extfig{#1}{\tikz{\node[font=\small,tight,minimum width=.24\columnwidth]{#1\phantom{g}};}}}
\pair{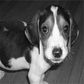}\
\pair{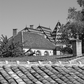}\\[2pt]
\pair{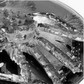}\
\pair{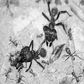}\\[2pt]
\pair{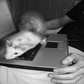}\
\pair{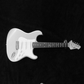}\\[2pt]
\pair{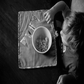}\
\pair{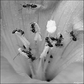}\\[2pt]
\lab{pooling}\ \lab{dense}\
\lab{pooling}\ \lab{dense}
\caption{Examples overlaid with correct \emph{class activation maps}~\cite{zhou2016} (red is high activation for ground truth) on Resnet-12 (\cf ~\secref{sec:exp}) trained with global average pooling or \classification (\cf~\eq{dc-map}). From top to bottom: base classes, classified correctly by both (walker hound, tile roof); novel classes, classified correctly by both (king crab, ant); novel classes, \classification is better (ferret, electric guitar); novel classes, pooling is better (mixing bowl, ant). In all cases, \classification results in smoother activation maps that are more aligned with objects.}
\label{fig:saliency}
\vspace{-2mm}
\end{figure}

As discussed in ~\secref{sec:background}, the \emph{embedding network} $\phi_\theta: \cX \to \real^{r \times d}$ maps the input to an embedding that is a tensor. There are two common ways of handling this high-dimensional representation, as illustrated in Figure~\ref{fig:dc}.

The first is to apply one or more fully connected layers, for instance in
networks C64F, C128F in few-shot learning~\cite{vinyals2016,snell2017,gidaris2018}. This can be seen as \emph{flattening} the activation into a long vector and multiplying with a weight vector of the same length per class; alternatively, the weight parameter is a tensor of the same dimension as the embedding. This representation is discriminative, but not invariant.

The second way is to apply \emph{global pooling} and reduce the embedding into a smaller vector of length $d$, for instance in
small ResNet architectures used more recently in few-shot learning~\cite{mishra2017,gidaris2018,oreshkin2018}. This reduces dimensionality significantly, so it makes sense if $d$ is large enough. It is an invariant representation, but less discriminative.

In this work we follow a different approach that we call \emph{\classification} and is illustrated in Figure~\ref{fig:dc3}. We view the embedding $\phi_\theta(\vx)$ as a collection of vectors $[\phi^{(k)}(\vx)]_{k=1}^r$, where $\phi^{(k)}(\vx) \in \real^d$ for $k \in [r]$\footnote{Given tensor $\va \in \real^{m \times n}$, denote by $\va^{(k)}$ the $k$-th $n$-dimensional slice along the first group of dimensions for $k \in [m]$.}. For a 2d image input and a convolutional network, $\phi_\theta(\vx)$ consists of the activations of the last convolutional layer, that is a tensor in $\real^{w \times h \times d}$ where $r = w \times h$ is its spatial resolution. Then, $\phi^{(k)}(\vx)$ is an embedding in $\real^d$ that represents a single spatial location $k$ on the tensor.

When learning from the training data $(X,Y)$ over base classes $C$ (stage 1), we adopt the simple approach of training a \emph{parametric linear classifier} on top of the embedding function $\phi_\theta$, like~\cite{qi2018} and the initial training of~\cite{gidaris2018}. The main difference in our case is that the weight parameters do \emph{not} have the same dimensions as $\phi_\theta(\vx)$; they are rather vectors in $\real^d$ and they are \emph{shared} over all spatial locations. More formally, let $\vw_j \in \real^d$ be the weight parameter of class $j$ for $j \in C$. Then, similarly to~\eq{print-map}, the classifier mapping
$f_{\theta,W}: \cX \to \real^{r \times c}$ is defined by
\vspace{-0.1cm}
\begin{align}
	f_{\theta,W}(\vx) \defn \left[
		\vsigma \left( [s_\tau(\phi_\theta^{(k)}(\vx), \vw_j)]_{j=1}^c \right)
	\right]_{k=1}^r
\label{eq:dc-map}
\end{align}
for $\vx \in \cX$, where
$W \defn (\vw_1,\dots,\vw_c)$ is the collection of class weights and $s_\tau$ is the \emph{scaled cosine similarity} defined by~\eq{cosine}, with $\tau$ being a learnable parameter as in~\cite{qi2018,gidaris2018}\footnote{\emph{Temperature scaling} is frequently encountered in various formats in several works to enable soft-labeling~\cite{hinton2015} or to improve cosine similarity in the final layer~\cite{wang2017b, oreshkin2018,gidaris2018, qi2018, hoffer2018}.}. Here $f_{\theta,W}(\vx)$ is a $r \times c$ tensor: index $k$ ranges over spatial resolution $[r]$ and $j$ over classes $[c]$.

\emph{This operation is a $1 \times 1$ convolution followed by depth-wise softmax}. Then, $f_{\theta,W}^{(k)}(\vx)$ at spatial location $k$ is a vector in $\real^c$ representing confidence over the $c$ classes. On the other hand, $f_{\theta,W}^{(:,j)}(\vx)$ is a vector in $\real^r$ representing confidence of class $j$ for $j \in [c]$ as a function of spatial location.\footnote{Given tensor $\va \in \real^{m \times n}$, denote by $\va^{(:,j)}$ the $j$-th $m$-dimensional slice along the second group of dimensions for $j \in [n]$.} For a 2d image input, $f_{\theta,W}^{(:,j)}(\vx)$ is like a \emph{class activation map} (CAM)~\cite{zhou2016} for class $j$, that is a 2d map roughly localizing the response to class $j$, but differs in that softmax suppresses all but the strongest responses at each location.

\begin{figure*}[!ht]
\centering
\includegraphics[width=0.95\linewidth]{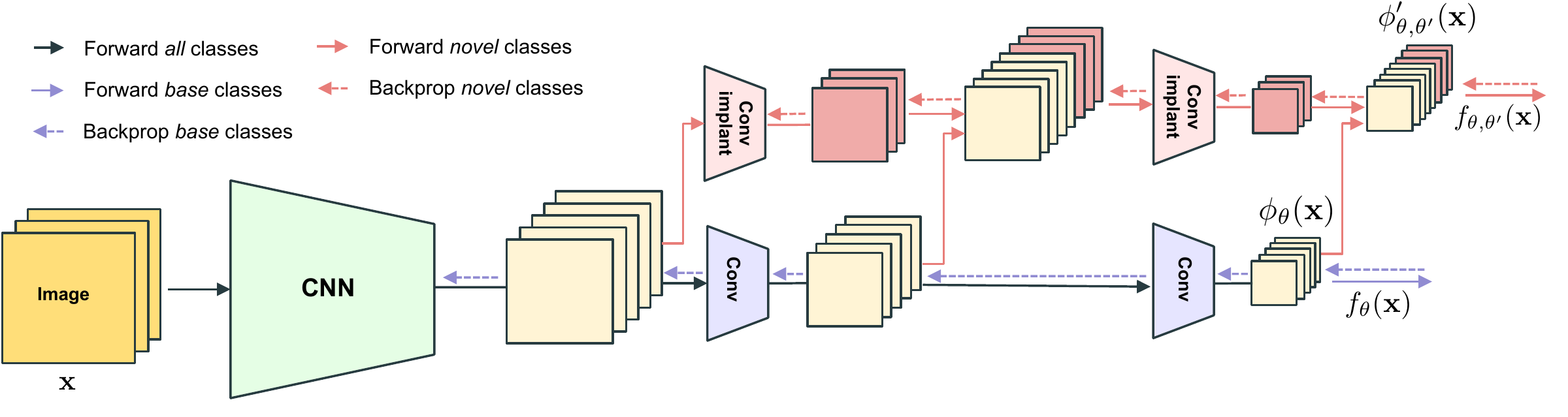}
\caption{\emph{Neural implants for CNNs.} The implants are convolutional filters operating in a new processing stream parallel to the base network. The input of an implant is the depth-wise concatenation of hidden states from both streams. When training neural implants, previously trained parameters are frozen. Purple and black arrows correspond to stage 1 flows; red and black to stage 2.}
\label{fig:implanting}
\end{figure*}

Given the definition~\eq{dc-map} of $f_{\theta,W}$, training amounts to minimizing over $\theta,W$ the \emph{cost function}
\vspace{-0.1cm}
\begin{align}
	J(X, Y; \theta, W) \defn \sum_{i=1}^n \sum_{k=1}^r \ell(f_{\theta,W}^{(k)}(\vx_i), y_i),
\label{eq:dc-cost}
\end{align}
where $\ell$ is cross-entropy~\eq{ce}. \emph{The loss function applies to all spatial locations and therefore the classifier is encouraged to make correct predictions
everywhere}.

Learning a new task with support data $(\nX,\nY)$ over novel classes $\nC$ (stage 2) and inference are discussed in~\secref{sec:stage2} and~\secref{sec:classify} respectively.

\head{Discussion.} The same situation arises in \emph{semantic segmentation}~\cite{long2015,noh2015}, where given per-pixel labels, the loss function applies per pixel and the network learns to make localized predictions on upsampled feature maps rather than just classify. In our case there is just one image-level label and the low resolution, \eg $5 \times 5$, of few-shot learning settings allows us to assume that the label applies to all locations due to large receptive field.

\Classification improves the spatial distribution of class activations, as shown in Figure~\ref{fig:saliency}. By encouraging all spatial locations to be classified correctly, we are encouraging the embedding network to identify all parts of the object of interest rather than just the most discriminative details. Since each location on a feature map corresponds to a region in the image where only part of the object may be visible, our model behaves like \emph{implicit data augmentation} of exhaustive shifts and crops over a dense grid with a single forward pass of each example in the network.

\subsection{Implanting}
\label{sec:implant}

From the learning on the training data $(X,Y)$ of base classes $C$ (stage 1) we only keep the embedding network $\phi_\theta$ and we discard the classification layer. The assumption is that features learned on base classes are generic enough to be used for other classes, at least for the bottom layers~\cite{yosinski2014}. However, given a new few-shot task on novel classes $\nC$ (stage 2), we argue that we can take advantage of the support data $(\nX,\nY)$ to find new features that are discriminative for the task at hand, at least in the top layers.

\subsubsection{Architecture}
\label{sec:arch}

We begin with the embedding network $\phi_\theta$, which we call \emph{base network}. We widen this network by adding new convolution kernels in a number of its top convolutional layers. We call these new neurons \emph{implants}. While learning the implants, we keep the base network parameters frozen, which preserves the representation of the base classes.

Let $\va_l$ denote the output activation of the convolutional layer $l$ in the base network. The implant for this layer, if it exists, is a distinct convolutional layer with output activation $\nva_l$. Then the input of an implant at the next layer $l+1$ is the depth-wise concatenation $[\va_l, \nva_l]$ if $\nva_l$ exists, and just $\va_l$ otherwise. If $\new{\theta_l}$ are the parameters of the $l$-th implant, then we denote by $\ntheta \defn (\new{\theta_{l_0}},\dots,\new{\theta_L})$ the set of all new parameters, where $l_0$ is the first layer with an implant and $L$ the network depth. The \emph{widened} embedding network is denoted by $\phi_{\theta,\ntheta}$.

As illustrated in Figure \ref{fig:implanting}, we are creating a new stream of data in parallel to the base network. The implant stream is connected to the base stream at multiple top layers and leverages the previously learned features by learning additional connections for the new tasks.

\vspace{-0.1cm}
\head{Why implanting?}
In several few-shot learning works, in particular metric learning, it is common to focus on the top layer of the network and learn or generate a new classifier for the novel classes. The reason behind this choice underpins a major challenge in few-shot learning: deep neural networks are prone to overfitting. With implanting, we attempt to diminish this risk by adding a limited amount of new parameters, while preserving the previously trained ones intact. Useful visual representations and parameters learned from base classes can be quickly squashed during fine-tuning on the novel classes. With implants, we \emph{freeze} them and train only the new neurons added to the network, maximizing the contribution of the knowledge acquired from base classes.
\vspace{-0.3cm}
\subsubsection{Training}
\label{sec:stage2}

To learn the implants only makes sense when a new task is given with support data $(\nX,\nY)$ over novel classes $\nC$ (stage 2). Here we use an approach similar to prototypical networks~\cite{snell2017} in the sense that we generate a number of fictitious \emph{subtasks} of the new task, the main difference being that we are now working on the novel classes.

We choose the simple approach of using each one of the given examples alone as a query in one subtask
while all the rest are used as support examples. This involves no sampling and the process is deterministic. Because only one example is missing from the true support examples, each subtask approximates the true task very well.

In particular, for each $i \in \nN \defn [\nn]$, we define a \emph{query set} $Q_i \defn \{i\}$ and a \emph{support set} $S_i \defn \nN \setminus Q_i$. We compute class prototypes $P_i$ on index set $S_i$ according to~\eq{proto}, where we replace $\phi_\theta$ by $\phi_{\theta,\ntheta}$ and $\ntheta$ are the implanted parameters. We define the \emph{widened} network function $f_{\theta,\ntheta}[P_i]$ on these prototypes by~\eq{proto-map}
with a similar replacement. We then freeze the base network parameters $\theta$ and train the implants $\ntheta$ by minimizing a cost function like~\eq{proto-cost}. Similarly to~\eq{proto-cost} and taking all subtasks into account, the overall cost function we are minimizing over $\ntheta$ is given by
\vspace{-0.3cm}
\begin{align}
	J(\nX, \nY; \theta, \ntheta) \defn
		\sum_{i=1}^{\nn} \ell(f_{\theta,\ntheta}[P_i](\nvx_i), \ny_i),
\label{eq:s2-cost}
\end{align}
where $\ell$ is cross-entropy~\eq{ce}.

In~\eq{s2-cost}, activations are assumed flattened or globally pooled. Alternatively, we can \emph{\classify} them and apply the loss function to all spatial locations independently. Combining with \eq{dc-cost}, the cost function in this case is
\vspace{-0.3cm}
\begin{align}
	J(\nX, \nY; \theta, \ntheta) \defn
		\sum_{i=1}^{\nn} \sum_{k=1}^r \ell(f_{\theta,\ntheta}^{(k)}[P_i](\nvx_i), \ny_i).
\label{eq:s2-dc-cost}
\end{align}
Prototypes in~\eq{s2-cost} or~\eq{s2-dc-cost} are recomputed at each iteration based on the current version of implants. Note that this training setup does not apply to the 1-shot scenario as it requires at least two \emph{support} samples per class.
\subsection{Inference on novel classes}
\label{sec:classify}

Inference is the same whether the embedding network has been implanted or not. Here we adopt the prototypical network model too. What we have found to work best is to perform global pooling of the embeddings of the support examples and compute class prototypes $P \defn (\vp_1,\dots,\vp_{\nc})$  by~\eq{proto}. Given a query $\vx \in \cX$, the standard prediction is then to assign it to the nearest prototype
\begin{align}
	\arg\max_{j \in \nC} s(\phi_{\theta,\ntheta}(\vx), \vp_j),
\label{eq:pred}
\end{align}
where $s$ is cosine similarity~\cite{snell2017}. Alternatively, we can \emph{\classify} the embedding $\phi_{\theta,\ntheta}(\vx)$, soft-assigning independently the embedding $\phi_{\theta,\ntheta}^{(k)}(\vx)$ of each spatial location, then average over all locations $k \in [r]$ according to
\begin{align}
	f_{\theta,\ntheta}[P](\vx) \defn
		\frac{1}{r} \sum_{k=1}^r \vsigma \left(
			[s_\tau(\phi_{\theta,\ntheta}^{(k)}(\vx), \vp_j)]_{j=1}^{\nc}
		\right),
\label{eq:pred-dc}
\end{align}
where $s_\tau$ is the scaled cosine similarity~\eq{cosine}, and finally classify to
$\arg\max_{j \in \nC} f_{\theta,\ntheta}^j[P](\vx)$.

\section{Related work}
\label{sec:related}

\head{Metric learning} is common in few-shot learning. Multiple improvements of the standard softmax and cross-entropy loss are proposed by~\cite{wen2016, liu2016, zheng2018, wan2018, hariharan2017} to this end. Traditional methods like \emph{siamese networks} are also considered~\cite{chopra2005, schroff2015, koch2015} along with models that learn by comparing multiple samples at once~\cite{vinyals2016, sung2018, snell2017}. Learning to generate new samples~\cite{hariharan2017} is another direction. Our solution is related to \emph{prototypical networks}~\cite{snell2017} and \emph{matching networks}~\cite{vinyals2016}
but we rather use a parametric classifier.

\head{Meta-learning} is the basis of a large portion of the few-shot learning literature. Recent approaches can be roughly classified as: \emph{optimization-based methods}, that learn to initialize the parameters of a learner such that it becomes faster to fine-tune~\cite{finn2017, mishra2018, nichol2018}; \emph{memory-based methods} leveraging memory modules to store training samples or to encode adaptation algorithms~\cite{santoro2016, ravi2017}; \emph{data generation methods} that learn to generate new samples~\cite{wang2018}; \emph{parameter generating methods} that learn to generate the weights of a classifier~\cite{gidaris2018, qiao2018} or the parameters of a network with multiple layers~\cite{bertinetto2016, ha2017, wang2017, han2018}. The motivation behind the latter is that it should be easier to generate new parameters rather than to fine-tune a large network or to train a new classifier from scratch. By generating a single linear layer at the end of the network~\cite{gidaris2018, qi2018, qiao2018}, one neglects useful coarse visual information found in intermediate layers. We plug our neural implants at multiple depth levels, taking advantage of such features during fine-tuning and learning new ones.

\head{Network adaptation} is common when learning a new task or new domain. One solution is to learn to \emph{mask} part of the network, keeping useful neurons and re-training/fine-tuning the remaining neurons on the new-task~\cite{mallya2018b, mallya2018a}. Rusu \etal~\cite{rusu2016} rather \emph{widen} the network by adding new neurons in parallel to the old ones at every layer. New neurons receive data from all hidden states, while previously generated weights are frozen when training for the new task. Our neural implants are related to \cite{rusu2016} as we add new neurons in parallel and freeze the old ones. Unlike \cite{rusu2016}, we focus on low-data regimes, keeping the number of new implanted neurons small to diminish overfitting risks and train faster, and adding them only at top layers, taking advantage of generic visual features from bottom layers~\cite{yosinski2014}.

\section{Experiments}
\label{sec:exp}

We evaluate our method extensively on the \emph{mini}ImageNet and FC100 datasets. We describe the experimental setup and report results below.

\subsection{Experimental setup}

\head{Networks.}
In most experiments we use a ResNet-12 network~\cite{oreshkin2018} as our embedding network, composed of four residual blocks~\cite{he2016}, each having three 3$\times$3 convolutional layers with batch normalization~\cite{ioffe2015} and swish-1 activation function~\cite{ramachandran2017}. Each block is followed by 2$\times$2 max-pooling. The shortcut connections have a convolutional layer to adapt to the right number of channels. The first block has 64 channels, which is doubled at each subsequent block such that the output has depth 512. We also test \classification on a lighter network C128F~\cite{gidaris2018} composed of four convolutional layers, the first (last) two having 64 (128) channels, each followed by 2$\times$2 max-pooling.

\begin{table}[t]
\begin{center}

\footnotesize
\begin{tabular}{@{\msp}l@{\msp}@{\msp}c@{\msp}|@{\msp}c@{\msp}@{\msp}c@{\msp}@{\msp}c@{\msp}}
\toprule
Network & Pooling & 1-shot & 5-shot & 10-shot \\
\midrule
C128F &  \AVGpoolshort & 54.28 $\pm$0.18 & 71.60 $\pm$0.13 & 76.92 $\pm$0.12\\
C128F & \short & 49.84 $\pm$0.18 & 69.64 $\pm$0.15 & 74.61 $\pm$0.13\\
ResNet-12 & \AVGpoolshort & 58.61 $\pm$0.18 & 76.40 $\pm$0.13 & 80.76 $\pm$0.11\\
ResNet-12 & \short & 61.26 $\pm$0.20 & 79.01 $\pm$0.13 & 83.04 $\pm$0.12\\
\bottomrule
\end{tabular}
\end{center}
\caption{\emph{Average 5-way accuracy on novel classes of \emph{mini}ImageNet, stage 1 only.} Pooling refers to stage 1 training. \AVGpoolshort: \aVGtraining; \short: \classification. At testing, we use global max-pooling on queries for models trained with \classification, and global average pooling otherwise.}
\label{tab:C128F}
\end{table}

\begin{table*}[ht]
\begin{center}
\footnotesize
\begin{tabular}{ @{\msp}l@{\msp}@{\msp}c@{\msp}|@{\msp}c@{\msp}@{\msp}c@{\msp}@{\msp}c@{\msp}@{\msp}c@{\msp}@{\msp}c@{\msp}@{\msp}c@{\msp}}
\toprule
    Stage 1 training & & \multicolumn{4}{c@{\msp}}{Support/query pooling at testing} \\
\midrule
                                       & Support $\rightarrow$  & \multicolumn{2}{c@{\msp}}{\MAXpoolshort}    & \multicolumn{2}{c@{\msp}}{\AVGpoolshort}  \\
      &  Queries $\rightarrow$                                  &  \MAXpoolshort & \short  & \AVGpoolshort    & \short      \\

\midrule
      \multirow{3}{*}{\AVGtraining}    &        Base classes       &  63.55 $\pm$0.20  & 77.17 $\pm$0.11  & 79.37 $\pm$0.09 & 77.15 $\pm$0.11  \\
                                       &         Novel classes    & 72.25 $\pm$0.13     & 70.71 $\pm$0.14 & 76.40 $\pm$0.13& 73.28 $\pm$0.14      \\
                                       &            Both classes   & 37.74 $\pm$0.07    & 38.65 $\pm$0.05  & 56.25 $\pm$0.10& 54.80 $\pm$0.09      \\

   \midrule
                                       &       Base classes   & 79.28 $\pm$0.10    & \textbf{80.67} $\pm$\textbf{0.10} & \textbf{80.61} $\pm$\textbf{0.10}& \textbf{80.70} $\pm$\textbf{0.10}          \\
      \Classification                         &     Novel classes       & \textbf{79.01} $\pm$\textbf{0.13}  & 77.93 $\pm$0.13 & 78.55 $\pm$0.13& \textbf{78.95} $\pm$\textbf{0.13}         \\
                                       &     Both classes       & 42.45 $\pm$0.07 & 57.98 $\pm$0.10 & 67.53 $\pm$0.10 & \textbf{67.78} $\pm$\textbf{0.10}                                 \\

\bottomrule
\end{tabular}
\caption{
\emph{Average 5-way 5-shot accuracy on base, novel and both classes of \emph{mini}ImageNet with ResNet-12, stage 1 only.} \MAXpoolshort: global max-pooling; \AVGpoolshort: global average pooling; \short: \classification. Bold: accuracies in the confidence interval of the best one.}
\label{tab:slicing}
\end{center}
\end{table*}

\begin{table}
\begin{center}
\footnotesize
\begin{tabular}{@{\msp}l@{\msp}@{\msp}c@{\msp}|@{\msp}c@{\msp}@{\msp}c@{\msp}@{\msp}c@{\msp}}

\toprule
\multicolumn{2}{c|@{\msp}}{Stage 2 training} & \multicolumn{3}{c@{\msp}}{Query pooling at testing} \\
\midrule
Support & Queries & \AVGpoolshort & \MAXpoolshort & \short \\

\midrule
\MAXpoolshort & \MAXpoolshort & 79.03 $\pm$ 0.19 & 78.92 $\pm$ 0.19 & 79.04 $\pm$ 0.19 \\
\MAXpoolshort & \short & 79.06 $\pm$ 0.19 & 79.37 $\pm$ 0.18 & 79.15 $\pm$ 0.19 \\
\AVGpoolshort & \AVGpoolshort & 79.62 $\pm$ 0.19 & 74.57 $\pm$ 0.22 & \textbf{79.77} $\pm$ \textbf{0.19} \\
\AVGpoolshort & \short & 79.56 $\pm$ 0.19 & 74.58 $\pm$ 0.22 & 79.52 $\pm$ 0.19 \\

\bottomrule
\end{tabular}

\end{center}
\caption{
\emph{Average 5-way 5-shot accuracy on novel classes of \emph{mini}ImageNet with ResNet-12 and \implanting in stage 2.} At testing, we use \AVGpoolshort for support examples. \MAXpoolshort: global max-pooling; \AVGpoolshort: global average pooling; \short: \classification.
\label{tab:implants}}
\end{table}

\head{Datasets.}
We use \textbf{\emph{mini}ImageNet~\cite{vinyals2016}}, a subset of ImageNet ILSVRC-12~\cite{russakovsky2014} of 60,000 images of resolution $84\times84$, uniformly distributed over 100 classes. We use the split proposed in~\cite{ravi2017}: $C=64$ classes for training, 16 for validation and 20 for testing.

We also use \textbf{FC100}, a few-shot version of CIFAR-100 recently proposed by Oreshkin \etal~\cite{oreshkin2018}. Similarily to \emph{mini}ImageNet, CIFAR-100~\cite{krizhevsky2009} has 100 classes of 600 images each, although the resolution is $32\times32$. The split is $C=60$ classes for training, 20 for validation and 20 for testing. Given that all classes are grouped into 20 super-classes, this split does not separate
super-classes: classes are more similar in each split and the semantic gap between base and novel classes is larger.

\head{Evaluation protocol.}
The training set $X$ comprises images of the base classes $C$. To generate the support set $\nX$ of a few-shot task on novel classes, we randomly sample $\nC$ classes from the validation or test set and from each class we sample $\kay$ images. We report the \emph{average accuracy} and the corresponding 95\% \emph{confidence interval} over a number of such tasks. More precisely, for all implanting experiments, we sample 5,000 few-shot tasks with 30 queries per class, while for all other experiments we sample 10,000 tasks. Using the same task sampling, we also consider few-shot tasks involving \emph{base classes} $C$, following the benchmark of~\cite{gidaris2018}. We sample a set of extra images from the base classes to form a test set for this evaluation, which is performed in two ways: independently of the novel classes $\nC$ and jointly on the union $C \cup \nC$. In the latter case, base prototypes learned at stage 1 are concatenated with novel prototypes~\cite{gidaris2018}.

\head{Implementation details.}
In stage 1, we train the embedding network for 8,000 (12,500) iterations with mini-batch size 200 (512) on \emph{mini}ImageNet (FC100). On \emph{mini}ImageNet, we use stochastic gradient descent with Nesterov momentum. On FC100, we rather use Adam optimizer~\cite{kingma2014}. We initialize the scale parameter at $\tau=10$ ($100$) on \emph{mini}ImageNet (FC100). For a given few-shot task in stage 2, the implants are learned over 50 epochs with AdamW optimizer~\cite{loshchilov2017} and scale fixed
at $\tau=10$.

\subsection{Results}

\head{Networks.}
In Table \ref{tab:C128F} we compare
ResNet-12 to C128F, with and without \classification. We observe that \classification improves classification accuracy on novel classes for ResNet-12, but it is detrimental for the small network. C128F is only 4 layers deep and the receptive field at the last layer is significantly smaller than the one of ResNet-12, which is 12 layers deep. It is thus likely that units from the last feature map correspond to non-object areas in the image. Regardless of the choice of using \classification or not, ResNet-12 has a large performance gap over C128F. For the following experiments, we use exclusively ResNet-12 as our embedding network.

\head{\Classification.}
To evaluate stage 1, we skip stage 2 and directly perform testing. In Table \ref{tab:slicing} we evaluate 5-way 5-shot classification on \emph{mini}ImageNet with \aVGtraining and \classification at stage 1 training, while exploring different pooling strategies at inference. We also tried using global max-pooling at stage 1 training and got similar results as with global average pooling. \Classification in stage 1 training outperforms \aVGtraining in all cases by a large margin. It also improves the ability of the network to integrate new classes without forgetting the base ones. Using \classification at testing as well, the accuracy on both classes is 67.78\%, outperforming the best result of 59.35\% reported by~\cite{gidaris2018}. At testing, \classification of the queries with global average pooling of the support samples is the best overall choice. One exception is global max-pooling on both the support and query samples, which gives the highest accuracy for new classes but the difference is insignificant.

\begin{table}[t]
\begin{center}

\footnotesize
\begin{tabular}{ @{\msp}l@{\msp}@{\msp}c@{\msp}@{\msp}c@{\msp}@{\msp}c@{\msp}@{\msp}c@{\msp}}

 \toprule
      Method                                     &  1-shot            & 5-shot              &   10-shot    & \\

\midrule
      \AVGpoolshort                                       & 58.61 $\pm$ 0.18   & 76.40 $\pm$ 0.13   & 80.76 $\pm$ 0.11\\
      \short (ours)                      & \textbf{62.53} $\pm$ \textbf{0.19}   & 78.95 $\pm$ 0.13   & 82.66 $\pm$ 0.11\\
      \short + WIDE                        &  61.73 $\pm$ 0.19 & 78.25 $\pm$ 0.14  & 82.03 $\pm$ 0.12 \\
      \short + \Implantingshort (ours)     &  - & \textbf{79.77} $\pm$ \textbf{0.19} & \textbf{83.83} $\pm$ \textbf{0.16}  \\

\midrule
      MAML~\cite{finn2017}                       & 48.70 $\pm$ 1.8   &    63.10 $\pm$ 0.9 &   -      \\
      PN~\cite{snell2017}                        & 49.42 $\pm$ 0.78   &    68.20 $\pm$ 0.66 &   -      \\
      Gidaris et al.~\cite{gidaris2018}          &  55.45 $\pm$ 0.7  & 73.00 $\pm$ 0.6    & - \\
      PN~\cite{oreshkin2018}                     &  56.50 $\pm$ 0.4    & 74.20 $\pm$ 0.2      & 78.60 $\pm$ 0.4 \\
      TADAM~\cite{oreshkin2018}                  &  58.50              & 76.70                 & 80.80  \\

\bottomrule
\end{tabular}

\end{center}
\caption{
\emph{Average 5-way accuracy on novel classes of \emph{mini}\-ImageNet}. The top part is our solutions and baselines, all on ResNet-12. \AVGpoolshort: \aVGtraining (stage 1); \short: \classification (stage 1); WIDE: last residual block widened by 16 channels (stage 1); \Implantingshort: \implanting (stage 2). In stage 2, we use \AVGpoolshort on both support and queries. At testing, we use \AVGpoolshort on support examples and \AVGpoolshort or \short on queries, depending on the choice of stage 1. The bottom part results are as reported in the literature. PN: Prototypical Network~\cite{snell2017}. MAML~\cite{finn2017} and PN~\cite{snell2017} use four-layer networks; while PN~\cite{oreshkin2018} and TADAM~\cite{oreshkin2018} use the same ResNet-12 as us. Gidaris et al.~\cite{gidaris2018} use a Residual network of comparable complexity to ours.
}
\label{tab:sota}

\end{table}

\begin{table}[t]
\begin{center}

\footnotesize
\begin{tabular}{ @{\msp}l@{\msp}@{\msp}c@{\msp}@{\msp}c@{\msp}@{\msp}c@{\msp}@{\msp}c@{\msp}}

 \toprule
Method & 1-shot & 5-shot & 10-shot \\

\midrule
\AVGpoolshort & 41.02 $\pm$ 0.17 & 56.63 $\pm$ 0.16 & 61.65 $ \pm$0.15 \\
\short (ours) & \textbf{42.04} $\pm$ \textbf{0.17} & 57.05 $\pm$ 0.16 & 61.91 $\pm$ 0.16 \\
\short + \Implantingshort (ours) & - & \textbf{57.63} $\pm$ \textbf{0.23} & \textbf{62.91} $\pm$ \textbf{0.22} \\

\midrule
PN~\cite{oreshkin2018} & 37.80 $\pm$ 0.40 & 53.30 $\pm$ 0.50 & 58.70 $\pm$ 0.40 \\
TADAM~\cite{oreshkin2018} & 40.10 $\pm$ 0.40 & 56.10 $\pm$ 0.40 & 61.60 $\pm$ 0.50 \\

\bottomrule
\end{tabular}

\end{center}
\caption{
\emph{Average 5-way accuracy on novel classes of FC100 with ResNet-12}. The top part is our solutions and baselines. \AVGpoolshort: \aVGtraining (stage 1); \short: \classification (stage 1); \Implantingshort: \implanting (stage 2). In stage 2, we use \AVGpoolshort on both support and queries. At testing, we use \AVGpoolshort on support examples and \AVGpoolshort or \short on queries, depending on the choice of stage 1. The bottom part results are as reported in the literature. All experiments use the same ResNet-12.
}
\label{tab:FC100}

\end{table}

\head{\Implanting.}
In stage 2, we add implants of 16 channels to all convolutional layers of the last residual block of our embedding network pretrained in stage 1 on the base classes with \classification. The implants are trained on the few examples of the novel classes and then used as an integral part of the widened embedding network $\phi_{\theta,\ntheta}$ at testing. In Table \ref{tab:implants}, we evaluate different pooling strategies for support examples and queries in stage 2. Average pooling on both is the best choice, which we keep in the following.

\head{Ablation study.} In the top part of Table \ref{tab:sota} we compare our best solutions with a number of baselines on 5-way \emph{mini}ImageNet classification. One baseline is the embedding network trained with \aVGtraining in stage 1. \Classification remains our best training option. In stage 2, the implants are able to further improve on the results of \classification. To illustrate that our gain does not come just from having more parameters and greater feature dimensionality, another baseline is to compare it to widening the last residual block of the network by 16 channels in stage 1. It turns out that such widening does not bring any improvement on novel classes. Similar conclusions can be drawn from the top part of Table~\ref{tab:FC100}, showing corresponding results on FC100. The difference between different solutions is less striking here. This may be attributed to the lower resolution of CIFAR-100, allowing for less gain from either dense classification or implanting, since there may be less features to learn.

\head{Comparison with the state-of-the-art.}
In the bottom part of Table~\ref{tab:sota} we compare our model with previous few-shot learning methods on the same 5-way \emph{mini}ImageNet classification. All our solutions outperform by a large margin other methods on 1, 5 and 10-shot classification. Our implanted network sets a new state-of-the art for 5-way 5-shot classification of \emph{mini}ImageNet. Note that prototypical network on ResNet-12~\cite{oreshkin2018} is already giving very competitive performance. TADAM~\cite{oreshkin2018} builds on top of this baseline to achieve the previous state of the art. In this work we rather use a cosine classifier in stage 1. This setting is our baseline GAP and is already giving similar performance to TADAM~\cite{oreshkin2018}. Dense classification and implanting are both able to improve on this baseline. Our best results are at least 3\% above TADAM~\cite{oreshkin2018} in all settings. Finally, in the bottom part of Table~\ref{tab:FC100} we compare our model on 5-way FC100 classification against prototypical network~\cite{oreshkin2018} and TADAM~\cite{oreshkin2018}. Our model
outperforms TADAM here too, though by a smaller margin.

\iffalse
\begin{figure}
\centering
\extfig{mnist}{
\begin{tikzpicture}
\begin{axis}[
    height=6cm,
    xlabel={$\ell_2$ distortion},
    ylabel={success probability},
    legend pos={south east},
]
    \addplot[red]   table{fig/eval/cwMnist.txt};   \leg{CW}
    \addplot[blue]  table{fig/eval/clipMnist.txt}; \leg{Ours}
    \addplot[green] table{fig/eval/fgsmMnist.txt}; \leg{FGSM}
    \addplot[brown] table{fig/eval/bimMnist.txt};  \leg{I-FGSM}
\end{axis}
\end{tikzpicture}
}
\caption[]{This is just a dummy figure to test \texttt{pgfplots}.}
\label{fig:p-l2-mnist}
\end{figure}
\fi

\section{Conclusion}
\label{sec:conclusion}

In this work we contribute to few-shot learning by building upon a simplified process for learning on the base classes using a standard parametric classifier. We investigate for the first time in few-shot learning the activation maps and devise a new way of handling spatial information by a dense classification loss that is applied to each spatial location independently, improving the spatial distribution of the activation and the performance on new tasks. It is important that the performance benefit comes with deeper network architectures and high-dimensional embeddings. We further adapt the network for new tasks by implanting neurons with limited new parameters and without changing the original embedding. Overall, this yields a simple architecture that outperforms previous methods by a large margin and sets a new state of the art on standard benchmarks.

{\small
\bibliographystyle{ieee}
\bibliography{main}
}

\end{document}